\documentclass[10pt,twocolumn,letterpaper]{article}

\usepackage{cvpr}
\usepackage{times}
\usepackage{epsfig}
\usepackage{graphicx}
\usepackage{amsmath}
\usepackage{amssymb}

\usepackage[pagebackref=false,breaklinks=true,letterpaper=true,colorlinks,bookmarks=false]{hyperref}

\cvprfinalcopy 

\def\cvprPaperID{****} 

\ifcvprfinal\fi
\begin{document}

\title{Vanishing Point Attracts Eye Movements in Scene Free-viewing}

\author{Ali Borji$^{\dagger}$ \ \ \ \ \ Mengyang Feng$^{+}$  \ \ \ \ \ Huchuan Lu$^{+}$ \\
$^{\dagger}$Computer Science Department, University of Wisconsin - Milwaukee, USA\\
$^{+}$Department of Electrical Engineering, Dalian University of Technology, China\\
{\url{borji@uwm.edu, archerfmy@163.com, luhuchuan@gmail.com}}
}

\maketitle

\begin{abstract}
\vspace{-10pt}
Eye movements are crucial in understanding complex scenes. 
By predicting where humans look in natural scenes, we can understand how they percieve scenes and priotriaze information for further high-level processing. Here, we study the effect of a particular type of scene structural information known as vanishing point and show that human gaze is attracted to vanishing point regions. We then build a combined model of traditional saliency and vanishing point channel that outperforms state of the art saliency models. 
\end{abstract}

\vspace{-15pt}
\section{Introduction}

Visual attention is one of the main components in scene understanding. 
Primates use eye movements to analyze complex natural scenes in real time. As an example, 
Yarbus (1967)~\cite{yarbus1967eye} showed that verbally-communicated task specification may dramatically cause 
highly variable spatio-temporal eye movements. As another example, Tanenhaus et al. (1995) ~\cite{tanenhaus1995integration} tracked fixations of subjects in an object manipulation task. 
They showed that visual context influenced syntactic processing when subjects received ambiguous verbal instructions. These two examples demonstrate an interplay between attention and scene understanding.

Several attention models have been proposed to find bottom-up salient regions by detecting regions that stand out from their surroundings~\cite{1998Itti,borji2013state,TriesmanGelade,GarciaDiazJOV,harel2007graph,hou2007saliency,zhang2013saliency,borji2013quantitative}. 
Several cues that attract attention and guide eye movements have already been discovered (e.g., color, texture, motion, text, face, object center-bias, scene center-bias, cultural cues, and gaze direction). Scene structural information such as scene gist, scene layout, horizontal line, depth, and openness influence eye movements as well as human scene categorization~\cite{torralba2006contextual}. Here, we systematicaly investigate the role of vanishing point (VP) and perspective on eye movements.

\begin{figure}[t]
\vspace{-10pt}
\begin{center}
\includegraphics[width=\linewidth]{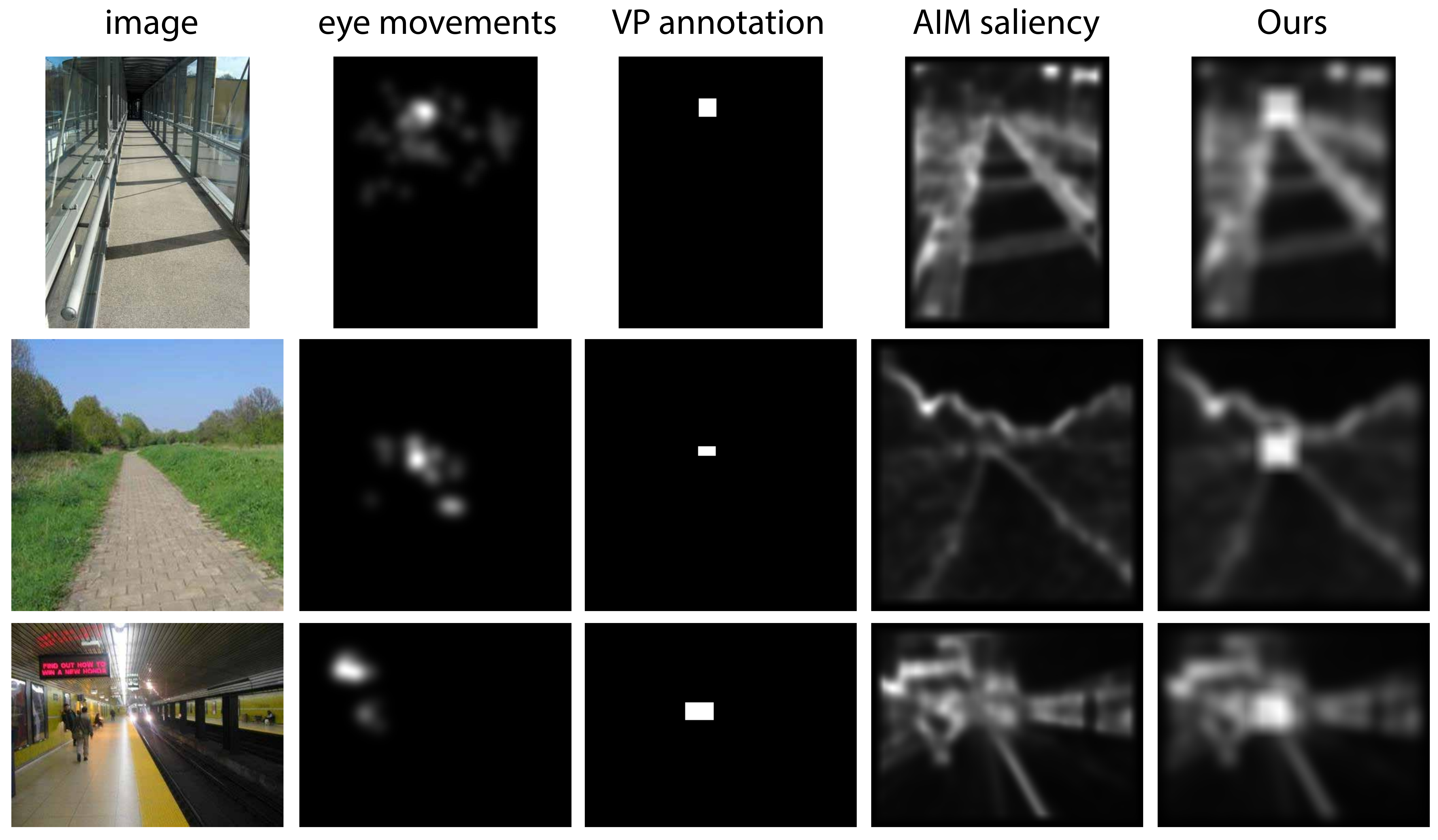}
\end{center}
\vspace{-10pt}
   \caption{Example stimuli and model prediction maps. We highlight the region surrounding vanishing point by a square. First two rows show success cases while the last row shows a failure. In this case, text is a more attractive cue than vanishing point.}
\label{fig:long}
\label{fig:onecol}
\vspace{-15pt}
\end{figure}

In graphical perspective, a vanishing point is a 2D point (in image plane) which is the intersection of parallel lines in the 3D world (but not parallel to the image plane). VP can be seen in fields, rail roads, streets, tunnels, forest, buildings, objects such as ladder, etc. It has been used in camera calibration, 3D reconstruction as well as in painting. Fig.~\ref{fig:onecol} shows some images with VPs, fixations and model outputs.

%

\section{Data}

Our stimuli are images with vanishing points from existing datasets (overall 115: 14 from DUT- OMRON~\cite{OMRON_DB} and 101 from MIT300~\cite{judd2012benchmark} dataset). These images have been shown shuffled among other images so subjects have had no bias in viewing them. We manually annotated VPs by drawing rectangles around them.








\section{Model}

\begin{figure}[t]
\begin{center}
\includegraphics[width=\linewidth]{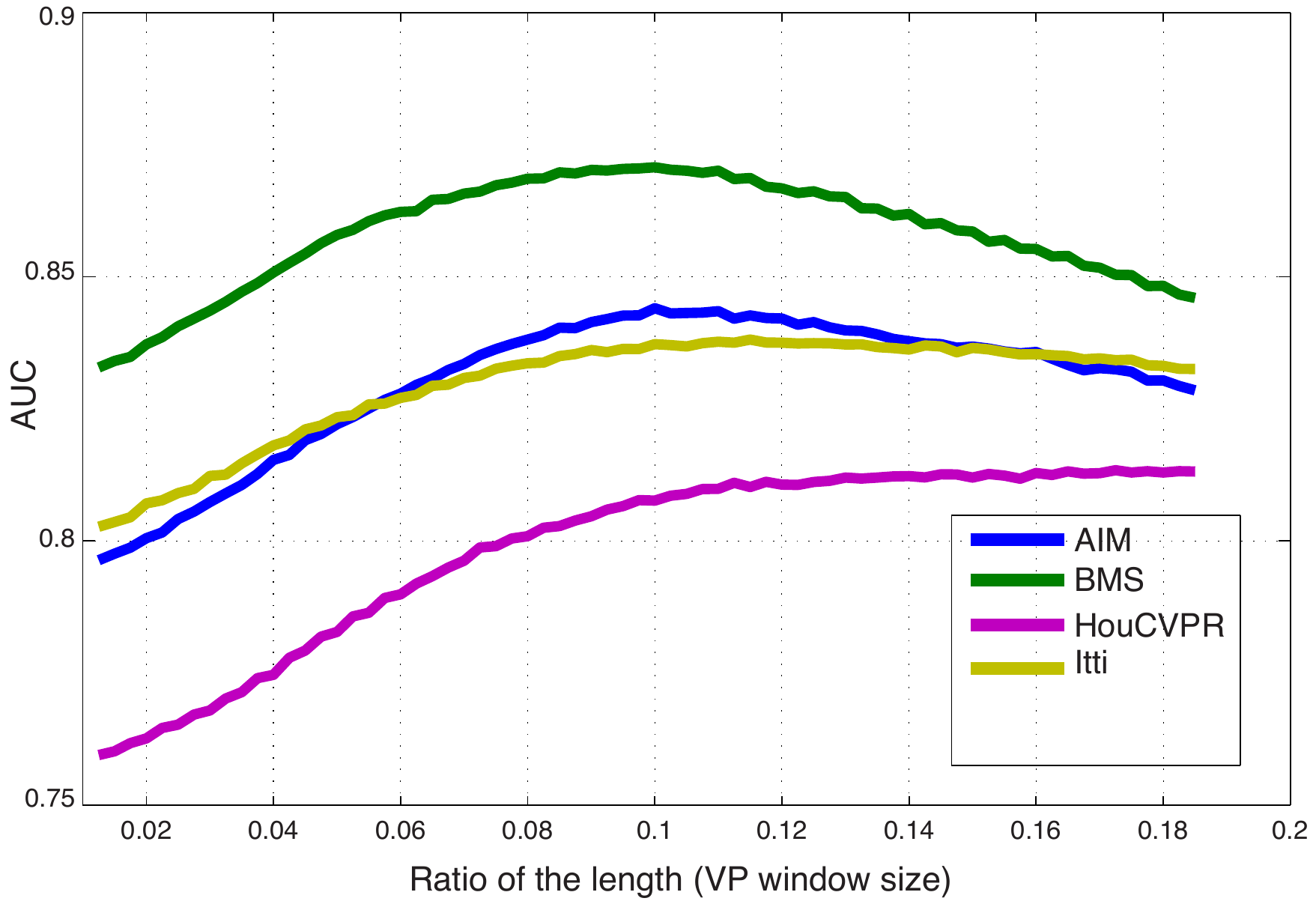}
\end{center}
\vspace{-15pt}
   \caption{Performance as a function of VP map size (Length ratio).}
\label{fig:long}
\label{fig:2}
\vspace{-15pt}
\end{figure}

Our model is the weighted linear combination of the saliency map and the vanishing point map (a square centered at the VP) as follows:
\begin{equation}
 SM =  \alpha \times S \ \ + \ \ (1-\alpha) \times VP, \ \ \alpha = 0 \ldots 1
\end{equation}

\noindent where S is the saliency map of the original model (one of 4 models: AIM, BMS, HouCVPR, Itti.), VP is the vanishing point map, and $\alpha$ is a parameter that controls the relative influence of the two maps. 
We empirically found that $\alpha=0.642$ leads to best results. The final map was smoothed by a Gaussian filter (see Fig.~\ref{fig:onecol}). We also optimized the size of the VP square (i.e., window size). Fig.~\ref{fig:2} shows performance of our model as a function of VP window size.



\section{Results}


We first show results over all data in Tables 1 and 2 using AUC and NSS~\cite{peters2005components}. As you can see, using both scores, we achieved a significant improvement over all baseline models (average of $\sim$\%6 improvement using AUC, $\sim$\%32 using NSS). VP only model performs well above chance but below baseline models. Our combined model shows $\sim$\%24 improvement over the VP model ($\sim$\%34 using NSS). Our best score is using BMS (AUC = 0.8707, NSS = 2.041).

\begin{table}
\begin{center}
\footnotesize{
\begin{tabular}{|l|c|c|c|c|c|c|}
\hline
Model &  & AIM & BMS & HouCVPR & Itti \\
\hline\hline
Oiginal & AUC & 0.7929 & 0.8323  & 0.7586 & 0.8009 \\
\hline
VP & AUC & 0.6653 & 0.6656  & 0.7154 & 0.6755 \\
\hline
Comb. & Best AUC & 0.8440 & 0.8707 & 0.8134 & .8381 \\
 	  & Ratio & 0.1000 & 0.1000 & 0.1725 & 0.1150 \\
	  & Window size & 40 & 40 & 69 & 46 \\
	  & Imp. vs orig. & 6.44\% & 4.61\% & 7.22\% & 4.64\% \\
	  & Imp. vs VP        & 26.86\% & 30.81\% & 13.70\% & 24.07\% \\
\hline
\end{tabular}
}
\end{center}
\vspace{-5pt}
\caption{AUC scores of original, VP, and combined models. The ratio is defined as $Ratio = 0.5*L/Max$, where $L$ is the length of VP square and $Max$ is the maximum side of the image (fixed to 400 px). Window size is size of the VP square (= L).}
\vspace{-5pt}
\end{table}

\begin{table}
\begin{center}
\footnotesize{
\begin{tabular}{|l|c|c|c|c|c|c|}
\hline
Model &  & AIM & BMS & HouCVPR & Itti \\
\hline\hline
Oiginal & NSS & 1.4075 & 1.5089  & 1.1764 & 1.3674 \\
\hline
VP & NSS & 1.3339 & 1.3443  & 1.3041 & 1.3136 \\
\hline
Comb. & Best NSS & 1.8112 & 2.0431 & 1.5762 & 1.6688 \\
 	  & Ratio & 0.0800 & 0.0650 & 0.0925 & 0.0900 \\
	  & Window size & 32 & 26 & 37 & 36 \\
	  & Imp. vs orig. & 28.68\% & 35.40\% & 39.99\% & 22.04\% \\
	  & Imp. vs VP        & 35.78\% & 51.98\% & 20.86\% & 27.04\% \\
\hline
\end{tabular}
}
\end{center}
\caption{NSS scores of original, VP, and combined models.}
\vspace{-10pt}
\end{table}



Next, we check the statistical significance of our results. We perform cross validation by randomly splitting data into two halves. We train our model (i.e., finding the best $\alpha$ using the learned best window size) on train set and apply it to the test set. We repeat this procedure 20 times and compare the means. Results of statistical tests using t-test are as follows. AUC score of the combined model using AIM model is 0.85 (std= 0.01) which is significantly higher than the original model (0.8, p=  3.14e-16) and the VP only model (0.64, p= 4.02e-22). AUC scores for the combined model using other models (BMS, HouCVPR, and Itti) in order are: 0.88, 0.82, 0.84 and all are significantly higher than the original and VP only models (p $\le$ 1e-17). The original model scores better than the VP only model and both perform significantly above chance (AUC = 0.5). We obtain the same pattern of results using NSS score. Using NSS, VP only model outperforms the original model. We obtain nearly the same best $\alpha$ using all original models.

\vspace{-5pt}
\section{Conclusion \& future work}
\vspace{-5pt}

We showed that vanishing point is an strong predictor of eye movements in free viewing by proposing a combined model. Since VP happens in many cases in real life when taking pictures, we believe that adding this channel to a model can in general enhance fixation prediction power.

We intend to study the followings in our future work: 1) Do people prioritize vanishing points in presence of other salient regions in a scene? We will design behavioral studies (using reaction time) to study this, 2) We intend to apply a auto detector for vanishing point to replace our annotations, 3) Investigate other forms of biasing vanishing point regions (e.g., using a circle or a Gaussian), and 4) Collecting a large dataset of eye movements on images with vanishing points.

\vspace{-5pt}
{\footnotesize
\bibliographystyle{ieee}
\bibliography{egbib}
}

\end{document}